# FTN: Foreground-Guided Texture-Focused Person Re-Identification

Donghaisheng Liu,[1][*] Shoudong Han,[1][*][†] Yang Chen,[1] Chenfei Xia,[1] Jun Zhao[2]

National Key Laboratory of Science and Technology on Multispectral Information Processing, School of Artificial Intelligence and Automation, Huazhong University of Science and Technology,[1] Nanyang Technological University[2]
{donghaisheng, shoudonghan, chenyang, chenfei_xia}@hust.edu.cn,[1] junzhao@ntu.edu.sg[2]

## Abstract

Person re-identification (Re-ID) is a challenging task as persons are often in different backgrounds. Most recent Re-ID methods treat the foreground and background information equally for person discriminative learning, but can easily lead to potential false alarm problems when different persons are in similar backgrounds or the same person is in different backgrounds. In this paper, we propose a Foreground-Guided Texture-Focused Network (FTN) for Re-ID, which can weaken the representation of unrelated background and highlight the attributes person-related in an end-to-end manner. FTN consists of a semantic encoder (S-Enc) and a compact foreground attention module (CFA) for Re-ID task, and a texture-focused decoder (TF-Dec) for reconstruction task. Particularly, we build a foreground-guided semi-supervised learning strategy for TF-Dec because the reconstructed ground-truths are only the inputs of FTN weighted by the Gaussian mask and the attention mask generated by CFA. Moreover, a new gradient loss is introduced to encourage the network to mine the texture consistency between the inputs and the reconstructed outputs. Our FTN is computationally efficient and extensive experiments on three commonly used datasets Market1501, CUHK03 and MSMT17 demonstrate that the proposed method performs favorably against the state-of-the-art methods.

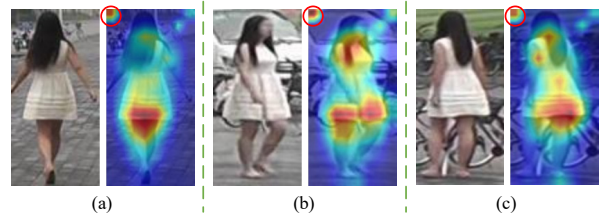

Figure 1: Visualization of false alarm of network caused by background interference. (a), (b), and (c) are the persons with the same identity and their activation map in different backgrounds. We can clearly see the high responses (but correspond to different attributes) of the background in the red circle.

## 1 Introduction

Person re-identification (Re-ID) aims to match persons with the same identity in different camera scenarios. It has significant application value in the field of intelligent video surveillance such as multi-object tracking (Bergmann, Meinhardt, and Lealtaixe 2019; P. Huang et al. 2020) and pedestrian retrieval (Han et al. 2019). Recently, with the development of deep learning, pedestrian Re-ID has received more and more attention from researchers, and considerable progress has been made. However, this task still faces many challenges, including differences in resolution, posture variations, background interference, noisy labels, occlusion, etc.

Among these challenges, in this paper, we focus on the background interference, which is often ignored but important.

For background interference challenge, some existing methods use the attention mechanism to obtain the potential semantic or structural correlations of persons on the basis of deep CNN to highlight the discriminative appearances information of human bodies (Chen, Deng, and Hu 2019; Chen et al. 2019; Dai et al. 2019). However, these attention-based methods, on the one hand, suffer from expensive computation, especially when the feature roots of tensors are calculated, which reduces the learning efficiency of the network. On the other hand, they do not perform explicit suppression processing against background interference. When an object in the background is considered to be a discriminative attribute of one person with high response, this attribute may not exist in other camera scenes (as shown in Figure 1), which will mislead the learning direction of the network.

In order to easily focus on person bodies to extract valuable foreground-related information, some methods (Guo et al. 2019; Miao et al. 2019; Wang et al. 2020) directly use image segmentation or pose estimation algorithms to determine the locations of body parts/key points for guiding the feature extractor generate body-related feature embeddings, which is effective in handling occlusion or posture variations. However, introducing an independent image segmentation/pose estimation module into the Re-ID task will complicate the construction of the overall model, and seriously affect the actual inference speed. More importantly, the locations of body parts/key points may be inaccurate due to background interference, which will result in great reduction of accuracy in Re-ID.

---

[*] Equal Contribution.
[†] Corresponding Author.

In this paper, we propose a Foreground-Guided Texture-Focused Network (FTN) to obtain discriminative person body information and suppress background semantic interference for an effective person Re-ID. FTN consists of a semantic encoder (S-Enc) and a compact foreground attention module (CFA) for Re-ID, and a texture-focused decoder (TF-Dec) for image reconstruction task. Our method is similar to (Jin et al. 2020) which builds a decoder (SA-Dec) for reconstructing the densely semantics aligned full texture image to achieve person semantics alignment. However, generating and processing the pseudo ground-truth corresponding to texture image is complicated and difficult, which limits the further application of (Jin et al. 2020) in real scenes. Unlike (Jin et al. 2020), to train and optimize TF-Dec, we do not use texture images that may have gaps with the real Re-ID dataset as ground-truths, but use mask-weighted original images to achieve TF-Dec foreground-guided semi-supervised learning which make the training conditions easy to meet and force S-Enc to weaken the false alarm effect caused by background interference. There are two sources of masks used for TF-Dec training, one is the general Gaussian mask, the other is the attention mask generated by our CFA. Note that FTN can be trained end-to-end because the mask generation is online, which is simpler than the method based on image segmentation or pose estimation, and has greater practical application value.

In particular, in order to improve the facticity of person texture information and the texture consistency captured by network between the input and reconstruction output, we introduce a novel gradient loss for TF-Dec. In addition, since CFA and TF-Dec are independent modules, FTN also has strong generality and extensibility, we can easily replace different S-Enc as needed. During inference, TF-Dec will be discarded to ensure the computationally efficient.

In summary, the contributions of our work mainly include:
- We propose a powerful framework FTN for person foreground-guided texture-focused representation learning to obtain robust and effective person Re-ID, which can be trained end-to-end.
- We build an independent simple TF-Dec trained by a novel gradient loss for person texture-focused reconstruction and can force FTN to be more discriminative.
- The lightweight CFA is designed to generate well-positioned attention mask for TF-Dec foreground-guided semi-supervised learning and can suppress the influence of background interference, which can be easily incorporated into other deep learning tasks.
- Experimental results show that the proposed FTN can achieve the state-of-the-art performance on Market1501 (Zheng et al. 2015), MSMT17 (Longhui Wei et al. 2018) and CUHK03 (Li et al. 2014).

## 2 Related Work

**Person Re-Identification.** Person Re-ID has made great progress in recent years based on deep neural networks (Zheng, Yang, and Hauptmann 2016). Most exiting Re-ID algorithms formulate their architectures inspired by object classification (Ahmed, Jones, and Marks 2015; T. Chen et al. 2019; Li et al. 2014; Sun et al. 2018; Z. Zheng, Zheng, and Yang 2017). To obtain discriminative feature representations, some focus on extracting the global feature of a person image (Chen et al. 2017; Qian et al. 2017; Yu et al. 2019; Zhou et al. 2019) while may not inhibit intra-class variations, e.g., occlusion, misalignment, background interference. Recently, some part-based methods are proposed to enhance the discriminative capabilities of person body parts, e.g. Sun Yifan et al. (2018) and Fu Yang et al. (2019) slice person images into horizontal grids for part learning and Hyunjong Park et al. (2020) consider relations between body parts for more discriminative feature extraction. In addition, some methods (Li, Zhu, and Gong 2018; Zhao et al. 2017) utilize attention mechanism to extract part-based spatial patterns from person images automatically. Nevertheless, these part-based methods do not treat background and foreground features differently, which will weaken the distinguishability of parts. There are also some methods (Chen et al. 2019; Chen et al. 2019; Wang et al. 2018; Xia et al. 2019; Xu et al. 2018) can obtain useful feature embeddings just rely on attention mechanism. However, the attention module may bring too much computational consumption, and attention-based methods still face the potential false alarm problem caused by background interference. Moreover, aligning the person body parts or feature embeddings is also a powerful method to deal with intra-class variations (Guo et al. 2019; Luo et al. 2019; Miao et al. 2019; Wang et al. 2020). However, these methods use image segmentation algorithms or pose estimation algorithms to achieve alignment, which will increase the difficulty of training, and in complex background scenes, their localization accuracy will be compromised. Different from the above methods, we design an end-to-end encoder-decoder network, which discriminately reconstructs the background and foreground through the texture-focused mechanism to achieve background interference suppression. Furthermore, we introduce the CFA module to further highlight the diverse discriminative features of foreground.

**Image Reconstruction for Person Re-ID.** To the best of our knowledge, image reconstruction (IR) is rarely used in person Re-ID as it is difficult to couple Re-ID and IR. Recent approaches which focus on deep learning based IR are designed to generate high resolution person images to address person Re-ID with various resolutions. Jiao Jiening et al. (2018) and Wang Zheng et al. (2018) recover some person visual details with the help of IR methods, and Mao Shunan et al. (2019) jointly optimize IR task and person feature extraction for resolution-invariant person Re-ID. However, in the real world, a person often does not have a corresponding high-resolution image, which limits the practical application of these methods. In order to ensure the efficiency of the Re-ID system, Jin et al. (2020) propose SAN which takes super-resolution (SR) descriptor as decoder to generate texture images and align person semantic information. However, generating and processing pseudo ground-truth texture images for SAN is complicated, which greatly increases the difficulty of training. Differently, our method does not require high-resolution clues for supervising, nor does it need to construct additional synthetic datasets. We design an independent efficient TF-Dec (does not participate in the inference stage which ensure the efficiency of inference) to achieve the foreground-guided texture-focused learning of person body and model the texture details of the image. Conveniently, the ground-truth used for TF-Dec training is only the original image processed by special masks.

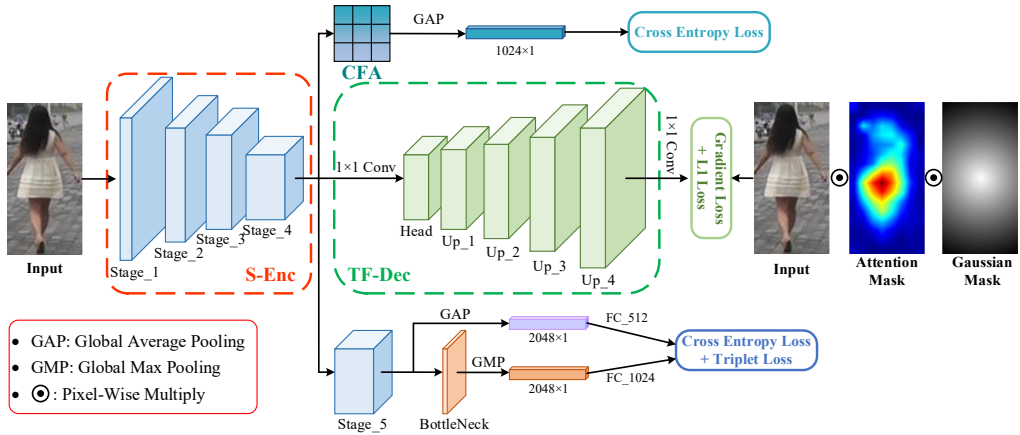

Figure 2: Proposed foreground-guided texture-focused network (FTN) for person Re-ID. Our model mainly consists of three components: a semantic encoder (S-Enc), a compact foreground attention module (CFA) and a texture-focused decoder (TF-Dec). Each stage of S-Enc corresponds to ResNet-50 and up blocks of TF-Dec are a series of upsampling modules with ratio 2. Attention mask and Gaussian mask are generated by CFA and Gaussian kernel respectively.

## 3 Methodology

### 3.1 Network Architecture Overview

The overall architecture of the proposed FTN is shown in Figure 2. FTN consists of a semantic encoder (S-Enc), a compact foreground attention module (CFA) for Re-ID and a texture-focused decoder (TF-Dec) for reconstruction.

S-Enc can be most common feature extraction backbones, such as ResNet (He et al. 2016), Densenet (Huang et al. 2017) and InceptionNet (Szegedy et al. 2016). Similar to most Re-ID algorithms, we use ResNet-50 by default to build an efficient S-Enc. We treat the original ResNet-50 down-sampling blocks as a series of stages, and use the output feature map of stage_4 as the high-dimensional input of CFA and TF-Dec, so that more sufficient body parts localization information can be modeled. The stage_5 is denoted as a global branch which is used to generate the identity sensitive feature embeddings. Referring to the previous work (Chen et al. 2019; Xia et al. 2019), we set the down-sampling ratio of stage_5 to 1 to obtain a larger feature map. In addition, as in (Xia et al. 2019), in order to build a more efficient feature extractor, we add a local branch (one bottleneck of ResNet-50) after stage_5.

CFA generates compact attention feature maps for persons in channel and position. Its detailed model structure will be described in Section 3.2, and the concept of *compact* will be further elaborated. Actually, there are two goals of CFA: For one thing, it is used as an important component to strengthen the ability of S-Enc to extract distinguishable features of person body; For another, in particular, it can provide foreground guidance to TF-Dec to focus on person texture reconstruction. As an independent module, CFA can also be easily incorporated into other vision tasks.

TF-Dec is a key and independent module of FTN, which contains head, body and tail (as shown in Figure 2). Our TF-Dec is lightweight that the number of channels of its head and body are always 64, and the number of channels of tail is 3 (RGB). The head contains $1 \times 1$ convolution, and the body is made up of 4 up blocks with upsampling ratio 2. Each up block contains PixelShuffle (Shi et al. 2016), $3 \times 3$ convolution and MSRB (Li et al. 2018) module. Unlike work (Jin et al. 2020) which directly utilizes the backbone feature layer to construct the upsampling blocks, our reconstruction will not conflict with the Re-ID. As mentioned, we use both the Gaussian mask and the attention mask weighted the input as the ground-truth of TF-Dec, so that TF-Dec can achieve semantic and texture reconstruction focusing on person-related attributes under the guidance of the foreground, which finally forces S-Enc to overcome background interference. Particularly, in order to improve the facticity of person texture information and the texture consistency captured by network between the input and reconstruction output, we propose a novel gradient loss (explained in Section 3.3) to train TF-Dec.

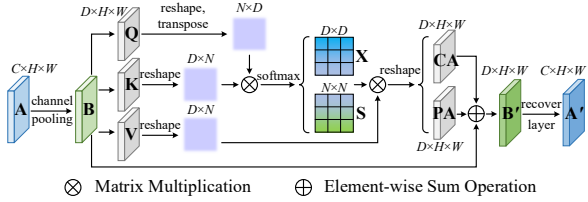

Figure 3: Compact Foreground Attention Module (CFA).

For Re-ID, as shown in Figure 2, we use global average pooling (GAP), global max pooling (GMP) and fully connected layers to obtain the feature vectors corresponding to CFA, global branch and local branch. During training, these feature vectors will be used to calculate the losses respectively, and during testing, they will be concatenated as the final feature embeddings.

### 3.2 Compact Foreground Attention Module

For person Re-ID, the attention module shows outstanding performance on highlighting the distinctive person features. In this paper, we propose a compact attention module that has fewer parameters, which is of practical significance for promoting the practical application of Re-ID.

Given a feature map, Chen Tianlong et al. (2019) designs two independent attention modules (we denote them as PAM-CAM) from two perspectives of channel and position, which have very similar structures. They first use a 3 × 3 convolution to reduce the number of channels, and then calculate the channel attention map and the position attention map respectively. After that, the attention maps and the input feature map are summed as the output, and the number of channels is restored by 3 × 3 convolution.

Inspired by (Chen 2019), we construct a compact foreground attention module (CFA), as shown in Figure 3. The improvements are: (1) We introduce the channel max pooling (CMP) to reduce the number of channels of input; (2) We share the feature maps used for generating channel attention map and position attention map; (3) The 1 × 1 convolution is used to restore the number of channels.

Compared with (Chen et al. 2019), our CFA is *compact*, which effectively reduces the amount of computation and parameters by using the smallest-scale 1 × 1 convolution kernel and sharing features. In particular, we introduce channel pooling to reduce the number of channels, which is different from most convolution-based channel reduction CNN models. Channel pooling can make the semantic features compact and discriminative without increasing the model parameters and directly eliminates some inactive features. Experiments show that for Re-ID, using channel pooling can achieve better retrieval performance than convolution.

In CFA, given the input feature maps **A** of size $C \times H \times W$, where C is the number of channels, H and W are the height and width respectively. We first perform 1 × $n$ channel max pooling on **A** to generate compact feature map **B** of size $D \times H \times W$, where $D = C / n$ (set $n = 2$ in experiments). Then **B** is fed into the 1 × 1 convolutional layer to produce feature maps **Q**, **K**, **V** of size $D \times H \times W$. **Q**, **K** is calculated after batch normalization, ReLU activation and softmax to obtain the channel affinity matrix **X** of size $D \times D$ and the position affinity matrix **S** of size $N \times N$, where $N = H \times W$. **X** and **S** perform matrix multiplication with **V** to generate channel attention map **CA** and position attention map **PA**. To further obtain the compact attention map **B′**, we have:

$$B'_i = \sum_{i=1}^{D}(\gamma CA_i + \phi PA_i + B_i), \quad (1)$$

where $\gamma$ and $\phi$ are the weights of **CA** and **PA** respectively, which are trainable as components of the model. Finally, the output attention map **A′** whose size is the same as **A** is generated through the recover layer with batch normalization and 1 × 1 convolution.

### 3.3 Gradient Loss for Texture-Focused Decoder

Many existing super-resolution algorithms use the L1 loss as the objective function of reconstruction learning (Dai et al. 2019; J. Li et al. 2018; Lim et al. 2017). L1 loss simply measures the differences between super-resolution reconstructed images and high-resolution images and is easy to calculate and implement. However, calculating the reconstruction loss directly from the pixel value will put the network in a huge searching space, and the network is not easy-optimized. Furthermore, we point out that the gradient of the image is also an important kind of feature that the super-resolution network should learn, and it reflects the changes of the texture. For this reason, in our method, we design a new gradient loss function for TF-Dec training to capture the rich texture details of persons and obtain texture consistency between input and output.

We calculate the horizontal gradient matrix $G^h$ and the vertical gradient matrix $G^v$ of the reconstructed image and the reconstruction ground-truth respectively, and use the L1 function to measure the differences between the reconstruct-

ed image gradient and the ground-truth gradient. Therefore, the gradient loss function can be defined as:

$$L_{gradient} = \frac{1}{b}\sum_{i=1}^{b}\left(\left\|G_i^{r,h} - G_i^{g,h}\right\|_1 + \left\|G_i^{r,v} - G_i^{g,v}\right\|_1\right), \quad (2)$$

$$G^h = \sum_{i=1}^{C}\sum_{y=1}^{H}\sum_{x=1}^{W-1}\left|I_i(x+1,y) - I_i(x,y)\right|^2, \quad (3)$$

$$G^v = \sum_{i=1}^{C}\sum_{y=1}^{H-1}\sum_{x=1}^{W}\left|I_i(x,y+1) - I_i(x,y)\right|^2, \quad (4)$$

where $b$ is the batch size, $r$ and $g$ respectively correspond to the reconstructed image and the ground-truth, $(x, y)$ represents the pixel coordinates. In this paper, we fix the reconstructed image and the ground-truth in the RGB color space, which means that C = 3.

### 3.4 Optimization and Training Scheme

**Optimization.** In this paper, we train our FTN end-to-end. For Re-ID, we use the cross-entropy loss $L_{CE}$ and the hard mining triplet loss (Hermans, Beyer, and Leibe 2017). Generally, the hard mining triplet loss is defined as:

$$L_{triplet} = \sum_{i=1}^{E}\sum_{a=1}^{M}[e + \overbrace{\max_{p=1...M}\left\|\mathbf{f}_a^{(i)} - \mathbf{f}_p^{(i)}\right\|_2}^{\text{hardest positive}} \\ - \underbrace{\min_{\substack{n=1...M \\ j=1...E, j\neq i}}\left\|\mathbf{f}_a^{(i)} - \mathbf{f}_n^{(j)}\right\|_2}_{\text{hardest negative}}]_+, \quad (5)$$

where $E$ and $M$ respectively represent the number of identities and instances in a mini-batch. $\mathbf{f}_a$, $\mathbf{f}_p$, $\mathbf{f}_n$ represent anchor, positive and negative feature vectors respectively and $e$ is the edge hyperparameter between intra-class distance and inter-class distance. In our experiments, we set $e = 0.3$.

As mentioned, we design a gradient loss function $L_{gradient}$ for TF-Dec training. But in order to accurately reconstruct the foreground texture-focused image, we also introduce the L1 loss on the basis of $L_{gradient}$. Thus, to train our FTN model, the final total loss is the weighted sum of the loss functions:

$$L_{FTN} = \lambda_1 L_{CE} + \lambda_2 L_{triplet} + \lambda_3 L_{gradient} + \lambda_4 L_{L1}, \quad (6)$$

where $\lambda_1, \lambda_2, \lambda_3, \lambda_4$ are the weighting factors.

**Training Scheme.** Similar to (Jin et al. 2020), FTN has two training strategies. One is to optimize the Re-ID task and the reconstruction task at the same time; the other is to alternately train the network related to Re-ID task and reconstruction task. Experiments show that using the second training strategy can make FTN get better optimization effect because there is no conflict between two tasks. Then we take every three batches as a group, set the $\lambda_1, \lambda_2, \lambda_3, \lambda_4$ of $L_{FTN}$ to 1, 0.1, 0, 0 for training in the first two epochs and set $\lambda_1, \lambda_2, \lambda_3, \lambda_4$ to 0, 0, 1, 1 for training in the third epoch.

## 4 Experiment

To demonstrate the effectiveness of FTN, we carry out comprehensive experiments on three public person Re-ID datasets: Market1501 (Zheng et al. 2015), MSMT17 (Wei et al. 2018) and CUHK03 (Li et al. 2014; Zhong et al. 2017). We first report a series of ablation study results to evaluate each component. Then, we compare the performance of FTN with exiting state-of-the-art methods. Finally, some analyses and visualized results are provided to show the fundamental role of FTN.

### 4.1 Implementation Details

During training, the input images are resized to 256×128 and then augmented by random horizontal flip, random crop, random path, and random erasing (Zhong et al. 2020). With the pretrained ResNet-50 on ImageNet, we finetune our model. In the warm-up stage (contains 51 epochs), we only train classifiers and attention modules. Subsequently, as described in Section 3.4, the network training related to Re-ID and reconstruction are performed alternately. We use Adam optimizer and set the initial learning rate to 3.5×10⁻⁴, then decays to 3.5×10⁻⁵, 3.5×10⁻⁶ and 3.5×10⁻⁷ after 150, 240 and 300 epochs. The total number of epochs is 360. Our training is done on two GTX 1080Ti GPUs based on PyTorch (Paszke et al. 2017) with batch size 32. Each batch contains 8 identities, with 4 instances per identity. For evaluation, we adopt the cumulative matching characteristics (CMC) at Rank-1, and the mean average precision (mAP) as evaluation metrics.

### 4.2 Ablation Study of FTN

We use the backbone with global and local branches as baseline, and perform ablation experiments on Market-1501 and CUHK03 (Labeled) datasets. In the ablation experiments, the training parameter settings of all control groups are the same.

**Effectiveness of CFA Module.** We first evaluate the effectiveness of the CFA module on FTN. We remove TF-Dec from TFN and calculate evaluation metrics by adding different types of CFA. As shown in Table 1, we observe that: (1) Using CFA can obtain a great improvement on both Market1501 and CUHK03 (Labeled) datasets; (2) The results related to **PA**, **CA** and CMP prove their effectiveness in forming CFA, and among them, **CA** has the greatest effect on the performance improvement of CFA. On Market1501, removing **CA** will bring a 1.1% decrease in Rank-1 accuracy to CFA; (3) Our CFA brings a more objective accuracy improvement than PAM-CAM (T. Chen et al. 2019), indicating that CFA has more advantages in highlighting valuable person clues.

We further compared the amount of computation and parameters of CFA and PAM-CAM (Chen et al. 2019), and the results show that our CFA is computationally and memory friendly with 0.168 GFLOPs and 1.3 million parameters, which reduce 83.4% calculation and parameters.
**Effectiveness of Different Reconstruction Guidance.** Then, we evaluate the effect of TF-Dec when the network has no CFA. Furthermore, based on the CFA, we analyze seven other semi-supervised learning strategies related to pedestrian texture-focused reconstruction.

| Model | Market-1501 | | CUHK03(L) | |
|---|---|---|---|---|
| | R-1 | mAP | R-1 | mAP |
| Base. | 94.5 | 85.8 | 67.6 | 66.0 |
| Base. + PAM-CAM (Chen et al. 2019) | 94.2 | 84.8 | 72.8 | 71.1 |
| Base. + CFA w/o **PA** | 95.5 | 87.8 | 74.0 | 72.7 |
| Base. + CFA w/o **CA** | 94.8 | 86.4 | 73.4 | 71.2 |
| Base. + CFA w/o CMP | 95.1 | 86.0 | 74.4 | 71.9 |
| Base. + CFA | **95.9** | **87.7** | **75.3** | **72.7** |

Table 1: Performance (%) comparisons of different attention modules on Market-1501 and CUHK03 (Labeled). **PA** and **CA**: Position and Channel Attention map. CMP: Channel Max Pooling. w/o: without. Base.: Baseline.

| Model | CFA | Gm | PAm | CAm | Market-1501 | | CUHK03(L) | |
|---|---|---|---|---|---|---|---|---|
| | | | | | R-1 | mAP | R-1 | mAP |
| Base. | – | – | – | – | 94.5 | 85.8 | 67.6 | 66.0 |
| FTN | × | √ | × | × | 95.7 | 87.1 | 73.3 | 71.2 |
| | √ | × | × | × | 95.2 | 87.5 | 76.4 | 75.0 |
| | √ | √ | × | × | 95.5 | 88.2 | 80.3 | 77.9 |
| | √ | √ | √ | × | 95.4 | 87.7 | 77.1 | 75.6 |
| | √ | √ | × | √ | 95.8 | 88.8 | 80.5 | **78.6** |
| | √ | √ | √ | √ | 95.6 | 87.7 | 78.6 | 77.3 |
| | √ | √ | √ | √ | **95.8** | **89.1** | **80.9** | 78.5 |

Table 2: Performance (%) comparisons between baseline (Base.) and FTN with different semi-supervision learning strategies of TF-Dec. These strategies have different reconstruction targets: (a) input weighted by the Gaussian mask (Gm) but without CFA module in the network; (b) just input; (c) input weighted by Gm; (d) input weighted by Gm and the position attention mask (PAm); (e) input weighted by Gm and the channel attention mask (CAm); (f) input weighted by Gm, PAm and CAm but without gradient loss; **(g)** input weighted by Gm, PAm and CAm (FTN-basic). The (a) to (g) correspond to the rows of results of FTN.

Table 2 presents the ablation study results and we can have the following conclusions: (1) From the results of (a), we can find that using only TF-Dec can improve the baseline on both datasets. It proves that texture-focused learning with foreground guidance is effective for improving the accuracy; (2) From (b) to **(g)**, using CFA and TF-Dec at the same time can bring further improvement to the baseline on both datasets. Especially when CFA generates attention mask for TF-Dec, the performance of the overall model has a significantly improvement; (3) No matter which reconstruction strategy is used, it can bring obvious improvement on the two datasets. When the reconstruction target is input weighted by the Gaussian mask and CFA mask, FTN can outperform baseline by a margin of 1.4% (Rank-1)/3.0% (mAP) on Market-1501 and 13.3% (Rank-1)/12.5% (mAP) on CUHK03 (Labeled). In addition, we notice that although PAm is effective, CAm is more effective. (4) From the last two rows in Table 2, we can see that $L_{gradient}$ has a significant improvement for Re-ID in terms of mining the textural patterns of input.

### 4.3 Comparison with State-of-the-Arts

Table 3 shows the performance comparison results between the proposed FTN and other state-of-the-art Re-ID methods on three datasets Market-1501, MSMT17 and CUHK03. To be fair, all methods do not use the re-ranking post-processing technique.

As shown in Table 3, compared with the baseline, our FTN can obtain the accuracy improvement by a large margin. Especially on CUHK03 dataset, whether we use labeled images or detected images, we have both more than 11% improvement in Rank-1 and mAP.

It can be seen that on all datasets, FTN can achieve competitive results with the recent state-of-the-art methods and is superior to most methods. Note that we obtain all results only on two GPUs. On Market-1501, FTN obtains 95.8% Rank-1 accuracy and 89.1% mAP, whose Rank-1 accuracy is quite close to SAN (Jin et al. 2020) and mAP obviously outperforms SAN. On CUHK03 (Labeled), FTN can still show outstanding performance, which is just no better than SONA$^{2+3}$-Net (Xia et al. 2019). On the two more difficult datasets CUHK03 (Detected) and MSMT17, FTN can achieve competitive performance as well.

As mentioned, our model architecture is similar to SAN. The decoder of SAN is used to achieve pedestrian semantic alignment, while FTN uses TF-Dec to achieve foreground-guided texture-focused reconstruction and force S-Enc and CFA to automatically eliminate false alarms caused by background interference for Re-ID. Compared with SAN that needs texture pseudo ground-truth for decoder training, our model is easier to train and has more practical value. From the results in Table 3, the performance of FTN is better than SAN, especially on CUHK03 and MSMT17.

### 4.4 Visualizations

**Visualization of Attention.** We first conduct a series of qualitative analyses on the attention module CFA. Figure 4 shows the response maps of three different pedestrians, which are generated by S-Enc's Stage_3 and CFA respectively. It can be found that as the middle feature layer of S-Enc, Stage_3's activation map has few feature response peaks. They are mostly related to attributes of persons but not comprehensive enough. In contrast, CFA's response map is more comprehensive and diverse, especially for some distinguishable person attributes (such as handbags). In fact, the attentiveness of CFA not only promotes S-Enc to extract more valuable clues, but also provides the possibility for TF-Dec to implement the texture-focused reconstruction of person-related attributes under the guidance of foreground.

| | Method | Market-1501 | | CUHK03 | | | | MSMT17 | |
|---|---|---|---|---|---|---|---|---|---|
| | | | | Labeled | | Detected | | | |
| | | R-1 | mAP | R-1 | mAP | R-1 | mAP | R-1 | mAP |
| Attention-based | AACN (Xu et al. 2018) | 85.9 | 66.9 | – | – | – | – | – | – |
| | HA-CNN (Li et al. 2018) | 91.2 | 75.7 | 44.4 | 41.0 | 41.7 | 38.6 | – | – |
| | Mancs (Wang et al. 2018) | 93.1 | 82.3 | 69.0 | 63.9 | 65.5 | 60.5 | – | – |
| | MHN-6(PCB) (Chen et al. 2019) | 95.1 | 85.0 | 77.2 | 72.4 | 71.7 | 65.4 | – | – |
| | SONA$^{2+3}$-Net (Xia et al. 2019) | 95.58 | 88.83 | 81.40 | 79.23 | 79.90 | 77.27 | – | – |
| | ABD-Net (Chen et al. 2019) | 95.60 | 88.28 | – | – | – | – | 82.30 | 60.80 |
| Others | AlignedReID++ (Luo et al. 2019) | 91.8 | 79.1 | – | – | 61.5 | 59.6 | 69.8 | 43.7 |
| | PCB+RPP (Sun et al. 2018) | 93.8 | 81.6 | 63.7 | 57.5 | – | – | 68.2 | 40.4 |
| | HPM (Fu et al. 2019) | 94.2 | 82.7 | 63.9 | 57.5 | – | – | – | – |
| | Relation-Net (Park and Ham 2020) | 95.2 | **88.9** | 77.9 | 75.6 | 74.4 | 69.6 | – | – |
| | OSNet (Zhou et al. 2019) | 94.8 | 84.9 | – | – | 72.3 | 67.8 | 78.7 | 52.9 |
| | $P^2$-Net (Guo et al. 2019) | 95.2 | 85.6 | 78.3 | 73.6 | 74.9 | 68.9 | – | – |
| | SAN (Jin et al. 2020) | 96.1 | 88.0 | 80.1 | 76.4 | **79.4** | 74.6 | 79.2 | 55.7 |
| | MGN+SIF (Long Wei et al. 2020) | 95.2 | 87.6 | 79.5 | 77.0 | 76.6 | 73.9 | – | – |
| Ours | Baseline | 94.5 | 85.8 | 67.6 | 66.0 | 64.4 | 62.1 | 76.5 | 52.5 |
| | FTN | **95.8** | 89.1 | **80.9** | **78.5** | 78.1 | **75.7** | **80.2** | **58.1** |

Table 3: Performance (%) comparisons with other state-of-the-art methods on Market-1501, CUHK03 and MSMT7 datasets. The best results are underlined, and the second best results are shown in bold.

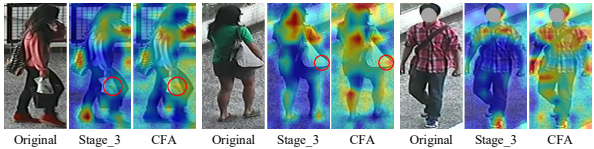

Figure 4: Visualization of activation maps generated by Stage_3 (from S-Enc) and CFA.

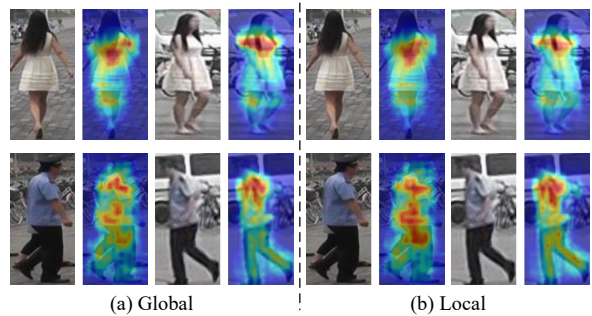

(a) Global    (b) Local

Figure 5: Visualization of activation maps generated by the global and local branches.

Although there are a small number of high response areas in the background, these activations will not cause false alarms for Re-ID, because we use both the attention mask and Gaussian mask to achieve foreground-focused learning and the Gaussian mask can suppress the saliency expression of the background in the attention mask.

**Visualization of Focused Area of S-Dec.** Finally, in order to verify whether S-Dec can eliminate the false alarm effect caused by background interference, we visualize the response maps of the global and local branches of S-Dec. As shown in Figure 5, it can be clearly found that for persons with the same identity but different backgrounds, whether it is a global branch or a local branch, there is a low response value in the background area, which means that S-Dec is not sensitive to background interference.

## 5  Conclusion

In this paper, we propose a Foreground-Guided Texture-Focused Network (FTN) to capture the guidance of foreground and generate the texture-focused features and further overcome the background interference for discriminative person Re-ID. With the semi-supervised learning strategy, FTN can be trained end-to-end conveniently. The visualizations intuitively demonstrate the foreground sensitivity and robustness of FTN. The ablation experiments show the effectiveness of each component in FTN, and the state-of-the-art performance can be obtained consistently on Market-1501, CUHK03 and MSMT17.

# 6 Acknowledgment

This work was supported by the National Natural Science Foundation of China under Grant No. 61105006; Open Fund of Key Laboratory of Image Processing and Intelligent Control (Huazhong University of Science and Technology), Ministry of Education under Grant No. IPIC2019-01; and the China Scholarship Council under Grant No. 201906165066.

# References

Ahmed, E.; Jones, M.; and Marks, T. K. 2015. An improved deep learning architecture for person re-identification. In *Proceedings of the IEEE Conference on Computer Vision and Pattern Recognition*, 3908-3916.

Bergmann, P.; Meinhardt, T.; and Lealtaixe, L. 2019. Tracking Without Bells and Whistles. In *Proceedings of the IEEE International Conference on Computer Vision*, 941-951.

Chen, B.; Deng, W.; and Hu, J. 2019. Mixed high-order attention network for person re-identification. In *Proceedings of the IEEE International Conference on Computer Vision*, 371-381.

Chen, T.; Ding, S.; Xie, J.; Yuan, Y.; Chen, W.; Yang, Y.; Ren, Z.; and Wang, Z. 2019. ABD-Net: Attentive but Diverse Person Re-Identification. In *Proceedings of the IEEE International Conference on Computer Vision*, 8351-8361.

Chen, W.; Chen, X.; Zhang, J.; and Huang, K. 2017. Beyond Triplet Loss: A Deep Quadruplet Network for Person Re-identification. In *Proceedings of the IEEE Conference on Computer Vision and Pattern Recognition*, 403-412.

Dai, T.; Cai, J.; Zhang, Y.; Xia, S.-T.; and Zhang, L. 2019. Second-order attention network for single image super-resolution. In *Proceedings of the IEEE Conference on Computer Vision and Pattern Recognition*, 11065-11074.

Fu, Y.; Wei, Y.; Zhou, Y.; Shi, H.; Huang, G.; Wang, X.; Yao, Z.; and Huang, T. 2019. Horizontal pyramid matching for person re-identification. In *Proceedings of the 33th AAAI Conference on Artificial Intelligence*, 8295-8302.

Guo, J.; Yuan, Y.; Huang, L.; Zhang, C.; Yao, J.; and Han, K. 2019. Beyond Human Parts: Dual Part-Aligned Representations for Person Re-Identification. In *Proceedings of the IEEE International Conference on Computer Vision*, 3642-3651.

Han, C.; Ye, J.; Zhong, Y.; Tan, X.; Zhang, C.; Gao, C.; and Sang, N. 2019. Re-ID Driven Localization Refinement for Person Search. In *Proceedings of the IEEE International Conference on Computer Vision*, 9814-9823.

He, K.; Zhang, X.; Ren, S.; and Sun, J. 2016. Deep residual learning for image recognition. In *Proceedings of the IEEE Conference on Computer Vision and Pattern Recognition*, 770-778.

Hermans, A.; Beyer, L.; and Leibe, B. 2017. In defense of the triplet loss for person re-identification. arXiv preprint. arXiv:1703.07737v4 [cs.CL]. Ithaca, NY: Cornell University Library.

Huang, G.; Liu, Z.; Van Der Maaten, L.; and Weinberger, K. Q. 2017. Densely connected convolutional networks. In *Proceedings of the IEEE Conference on Computer Vision and Pattern Recognition*, 4700-4708.

Huang, P.; Han, S.; Zhao, J.; Liu, D.; Wang, H.; and Yu, E. 2020. Refinements in Motion and Appearance for Online Multi-Object Tracking. arXiv preprint. arXiv:2003.07177v2 [cs.CL]. Ithaca, NY: Cornell University Library.

Jiao, J.; Zheng, W.-S.; Wu, A.; Zhu, X.; and Gong, S. 2018. Deep low-resolution person re-identification. In *Proceedings of the 32th AAAI Conference on Artificial Intelligence*, 6967-6974.

Jin, X.; Lan, C.; Zeng, W.; Wei, G.; and Chen, Z. 2020. Semantics-Aligned Representation Learning for Person Re-Identification. In *Proceedings of the 34th AAAI Conference on Artificial Intelligence*, 11173-11180.

Li, J.; Fang, F.; Mei, K.; and Zhang, G. 2018. Multi-scale residual network for image super-resolution. In *Proceedings of the European Conference on Computer Vision*, 517-532.

Li, W.; Zhao, R.; Xiao, T.; and Wang, X. 2014. Deepreid: Deep filter pairing neural network for person re-identification. In *Proceedings of the IEEE Conference on Computer Vision and Pattern Recognition*, 152-159.

Li, W.; Zhu, X.; and Gong, S. 2018. Harmonious attention network for person re-identification. In *Proceedings of the IEEE Conference on Computer Vision and Pattern Recognition*, 2285-2294.

Lim, B.; Son, S.; Kim, H.; Nah, S.; and Mu Lee, K. 2017. Enhanced deep residual networks for single image super-resolution. In *Proceedings of the IEEE Conference on Computer Vision and Pattern Recognition*, 136-144.

Luo, H.; Jiang, W.; Zhang, X.; Fan, X.; Qian, J.; and Zhang, C. 2019. AlignedReID++: Dynamically matching local information for person re-identification. *Pattern Recognition* 94: 53-61. doi.org/10.1016/j.patcog.2019.05.028.

Mao, S.; Zhang, S.; and Yang, M. 2019. Resolution-invariant Person Re-Identification. In *Proceedings of the Twenty-Eighth International Joint Conference on Artificial Intelligence*, 883-889.

Miao, J.; Wu, Y.; Liu, P.; Ding, Y.; and Yang, Y. 2019. Pose-guided feature alignment for occluded person re-identification. In *Proceedings of the IEEE International Conference on Computer Vision*, 542-551.

Park, H., Ham, B. 2020. Relation Network for Person Re-identification. In *Proceedings of the 34th AAAI Conference on Artificial Intelligence*, 11839-11847.

Paszke, A.; Gross, S.; Chintala, S.; Chanan, G.; Yang, E.; DeVito, Z.; Lin, Z.; Desmaison, A.; Antiga, L.; and Lerer, A. 2017. Automatic differentiation in pytorch.

Qian, X.; Fu, Y.; Jiang, Y.-G.; Xiang, T.; and Xue, X. 2017. Multi-scale deep learning architectures for person re-identification. In *Proceedings of the IEEE International Conference on Computer Vision*, 5399-5408.

Shi, W.; Caballero, J.; Huszár, F.; Totz, J.; Aitken, A. P.; Bishop, R.; Rueckert, D.; and Wang, Z. 2016. Real-time single image and video super-resolution using an efficient sub-pixel convolutional neural network. In *Proceedings of the IEEE Conference on Computer Vision and Pattern Recognition*, 1874-1883.

Sun, Y.; Zheng, L.; Yang, Y.; Tian, Q.; and Wang, S. 2018. Beyond part models: Person retrieval with refined part pooling (and a strong convolutional baseline). In *Proceedings of the European Conference on Computer Vision*, 480-496.

Szegedy, C.; Ioffe, S.; Vanhoucke, V.; and Alemi, A. A. 2016. Inception-v4, Inception-ResNet and the Impact of Residual Connections on Learning. In *Proceedings of the 30th AAAI Conference on Artificial Intelligence*, 4278-4284.

Wang, C.; Zhang, Q.; Huang, C.; Liu, W.; and Wang, X. 2018. Mancs: A multi-task attentional network with curriculum sampling for person re-identification. In *Proceedings of the European Conference on Computer Vision*, 365-381.

Wang, G. a.; Yang, S.; Liu, H.; Wang, Z.; Yang, Y.; Wang, S.; Yu, G.; Zhou, E.; and Sun, J. 2020. High-Order Information Matters: Learning Relation and Topology for Occluded Person Re-Identification. In *Proceedings of the IEEE Conference on Computer Vision and Pattern Recognition*, 6449-6458.

Wang, Z.; Ye, M.; Yang, F.; Bai, X.; and Satoh, S. i. 2018. Cascaded SR-GAN for Scale-Adaptive Low Resolution Person Re-identification. In *Proceedings of the Twenty-Seven International Joint Conference on Artificial Intelligence*, 3891-3897.

Wei, L.; Wei, Z.; Jin, Z.; Yu, Z.; Huang, J.; Cai, D.; He, X.; and Hua, X.-S. 2020. SIF: Self-Inspirited Feature Learning for Person Re-Identification. *IEEE Transactions on Image Processing* 29: 4942-4951. doi.org/10.1109/TIP.2020.2975712.

Wei, L.; Zhang, S.; Gao, W.; and Tian, Q. 2018. Person transfer gan to bridge domain gap for person re-identification. In *Proceedings of the IEEE Conference on Computer Vision and Pattern Recognition*, 79-88.

Xia, B. N.; Gong, Y.; Zhang, Y.; and Poellabauer, C. 2019. Second-order non-local attention networks for person re-identification. In *Proceedings of the IEEE International Conference on Computer Vision*, 3760-3769.

Xu, J.; Zhao, R.; Zhu, F.; Wang, H.; and Ouyang, W. 2018. Attention-aware compositional network for person re-identification. In *Proceedings of the IEEE Conference on Computer Vision and Pattern Recognition*, 2119-2128.

Yu, T.; Li, D.; Yang, Y.; Hospedales, T. M.; and Xiang, T. 2019. Robust person re-identification by modelling feature uncertainty. In *Proceedings of the IEEE International Conference on Computer Vision*, 552-561.

Zhao, L.; Li, X.; Zhuang, Y.; and Wang, J. 2017. Deeply-learned part-aligned representations for person re-identification. In *Proceedings of the IEEE International Conference on Computer Vision*, 3219-3228.

Zheng, L.; Shen, L.; Tian, L.; Wang, S.; Wang, J.; and Tian, Q. 2015. Scalable person re-identification: A benchmark. In *Proceedings of the IEEE International Conference on Computer Vision*, 1116-1124.

Zheng, L.; Yang, Y.; and Hauptmann, A. G. 2016. Person re-identification: Past, present and future. arXiv preprint. arXiv:1610.02984v1 [cs.CL]. Ithaca, NY: Cornell University Library.

Zheng, Z.; Zheng, L.; and Yang, Y. 2017. A discriminatively learned cnn embedding for person reidentification. *ACM Transactions on Multimedia Computing, Communications, and Applications* 14(1): 1-20. doi.org/10.1145/3159171.

Zhong, Z.; Zheng, L.; Cao, D.; and Li, S. 2017. Re-ranking person re-identification with k-reciprocal encoding. In *Proceedings of the IEEE Conference on Computer Vision and Pattern Recognition*, 1318-1327.

Zhong, Z.; Zheng, L.; Kang, G.; Li, S.; and Yang, Y. 2020. Random Erasing Data Augmentation. In *Proceedings of the 34th AAAI Conference on Artificial Intelligence*, 13001-13008.

Zhou, K.; Yang, Y.; Cavallaro, A.; and Xiang, T. 2019. Omni-scale feature learning for person re-identification. In *Proceedings of the IEEE International Conference on Computer Vision*, 3702-3712.